# Consistency and Consensus Driven for Hesitant Fuzzy Linguistic Decision Making with Pairwise Comparisons

Peijia Ren[a], Zixu Liu[b,*], Wei-Guo Zhang[a], Xilan Wu[c]

[a] *School of Business Administration, South China University of Technology, Guangzhou 510640, China*

[b] *Crop Science Centre, National Institute of Agricultural Botany, Cambridge CB3 0LE, U.K.*

[c] *Alliance Manchester Business School, The University of Manchester, Manchester M15 6PB, U.K.*

E-mails: pjren@scut.edu.cn (Peijia Ren); zixuxilan@gmail.com (Zixu Liu); wgzhang@scut.edu.cn (Wei-Guo Zhang); laceysharon@163.com (Xilan Wu)

**Abstract**

Hesitant fuzzy linguistic preference relation (HFLPR) is of interest because it provides an efficient way for opinion expression under uncertainty. For enhancing the theory of group decision making (GDM) with HFLPR, the paper introduces a method for addressing the GDM based on consistency and consensus measurements, which involves (1) defining a hesitant fuzzy linguistic geometric consistency index (HFLGCI) and proposing an algorithm for consistency checking and inconsistency improving for HFLPR; (2) proposing a worst consensus index based on the minimum similarity measure between each individual HFLPR and the overall perfect HFLPR in order to build an algorithm for consensus reaching based on the acceptable HLFPRs. The convergence and monotonicity of the proposed two procedures is proved. Experiments are performed to investigate the critical values of the defined HFLGCI, and comparative analyses are conducted to show the effectiveness of the proposed method. A case concerning the performance evaluation of venture capital guiding funds is given to illustrate the applicability of the proposed method. As an application of our work, an online decision-making portal is finally provided for decision makers to utilize the proposed method to solve GDM with HFLPRs.

1. **Introduction**

Evolution and development in the global economy have heightened the need for group decision making (GDM), a procedure for optimal choice determination involving a group of decision makers (DMs). Since it solves problems with a collective intelligence that avoids the lack of individual cognition, the relevant research of GDM has experienced unprecedented growth in the past years (Dyer & Forman, 1992; Liu et al., 2017; Kim et al., 1998; Ren et al., 2017; Yu & Lai, 2011; Zhang, 2003). Preference relation, which depicts DMs' pairwise comparisons, is significant to be used in GDM. Because DMs usually display limitations of thinking during the decision-making process (Simon, 1956, 1984), portraying the judgments that fit DMs' realistic mind is an efficient way to solve decision-making problems.

Based on the fact that linguistic terms (LTs) (Zadeh, 1975) describe objects with conforming DMs' expression habit, linguistic preference relation (LPR) (Herrera et al, 1995) was further proposed to compare two objects by "better than", "slightly worse than", among other LTs. With the increasing diversity and complexity of decision-making problems, the limited information and human thinking fuzziness get more noticeable. Describing DMs' judgements more elastically is necessary for ensuring the reasonability of the outcomes. To address this point, hesitant fuzzy linguistic preference relation (HFLPR) (Zhu & Xu, 2014a) was introduced to represent DMs' preferences based on consecutive LTs (Rodríguez et al., 2012; Liao & Xu, 2015), which is a more general concept of LPR and can flexibly show DMs' comparative judgements. An example for illustrating this concept is given as: "the superior degree of Alternative A to Alternative B is between slightly better and better".

Recent trends in decision making under uncertainty have led to a proliferation study of HFLPRs. As for the research on the consistency of a HFLPR, Zhu and Xu (2014a) and Zhang and Wu (2014) defined its additive consistency and multiplicative consistency, and respectively established two algorithms for consistency checking and inconsistency improving. Through introducing the interval consistency index for a HFLPR, Li et al. (2018) discussed the related consistency index measurement. Wu et al. (2019a, 2020) constructed mathematical programming models for deriving the acceptable consistent HFLPR based on discussing the least common multiple expansion principle and consistency property of aggregator, respectively. Mi et al. (2019) introduced a framework of hesitant fuzzy linguistic AHP, which addresses multiplicative consistency definition, a linear programming model for prioritizations, a feedback algorithm for improving the consistency of HFLPR according to the perfect one.

Another focused research on HFLPRs is the GDM process. With the compatibility measure between the synthetic HFLPR and the ideal HFLPR, Gou et al. (2017) provided a model for weight determination and a process for GDM with HFLPRs. Based on defining a distance measure between two HFLPRs and a consensus index presented by the divergence between individuals and the group, Lin and Wang (2020) built a group consensus framework involving an automatic consensus reaching mechanism and an interactive consensus reaching mechanism. Ren et al. (2021) put forward a group consensus feedback model based on considering the modified extents of HFLPRs given by DMs. In addition, the group consensus has been addressed based on considering the consistency of HFLPRs to ensure the reasonability of the decision-making results. By mapping a 2-tuple linguistic (2TL) into an equivalent form, Wu and Xu (2016) presented an additive consistency index for HFLPR based on the distance between 2TL and the equivalent form, and then constructed a feedback mechanism with consensus to address the GDM. With the consistency measurement of HFLPR and ordinal numbers, the ordinal consensus was applied to build a consensus procedure (Tang & Liao, 2018). Furthermore, on the basis of discussing the worst consistency, additive consistency, multiplicative consistency of an HFLPR, various consensus models aiming to minimize the divergence between individuals and the group were proposed (Chen et al., 2020; Wu et al, 2019b; Zhang & Chen, 2019, 2020)

The research on GDM with HFLPRs has great potential in theory and practice due to the superiority of HFLPRs in portraying fuzziness under uncertainty. However, the current works have gaps: (1) they proposed the consistency measures for an HFLPR based on the additive transitivity, multiplicative transitivity, order transitivity, ignored some other important consistency measures such as mean random consistency, geometric consistency, etc., which limits the theoretical framework of decision making with HFLPRs; (2) they mostly studied the group consensus by measuring the divergence among all individuals and the group, the perspectives for the group consensus would be broadened to enhance its applicability; (3) they mostly ignored the issue of applications for the public to use or evaluate decision models, it is hard for users to catch the different programming languages about the GDM models designed by different researchers, even if the programming codes are available for them.

To deal with the above gaps, the paper aims to study the consistency and consensus-based decision support algorithms with HFLPRs, and relatively give a running framework for implementing the online decision support system, which has the contributions:

(1) it defines the hesitant fuzzy linguistic geometric consistency index (HFLGCI) through establishing the relationships among the numerical type of preference relations and the linguistic type of preference relations (asymmetric form), which provides another important consistency property for an HFLPR and fruits its consistency theory.

(2) it proposes a worst consensus index based on describing the minimum similarity between each individual HFLPR and the overall perfect HFLPR, and builds a feedback consistency and consensus-driven mechanism for GDM with HFLPRs, which ensures the group consensus better to be achieved and increases the compatibility of the decision-making results.

(3) it provides an online decision-making portal[1] for DMs to utilize the proposed algorithms to handle decision-making problems. With the designed pod-based microservice architecture and friendly user interface (UI), the proposed online portal (digital platform) can deploy, run and manage the multilingual consistency and consensus-based decision support algorithms to provide decision support services or tools for DMs.

The paper can be organized as follows: in Section 2, we review several kinds of preference relations and the geometric consistency index (GCI). Section 3 defines the concept of HFLGCI and the consensus measure of a group of HFLPRs, and provides an algorithm for GDM with HFLPRs based on acceptable consistency and consensus measures. In Section 4, we discuss the critical values for the HFLGCI and make some comparative analyses with the existing methods. Section 5 applies the proposed algorithm to evaluate the performance of venture capital guiding funds. The online portal involving two proposed algorithms is developed as an application of our work in Section 6. The paper ends up with conclusions in Section 7.

---

[1] http://34.92.80.18/ (username: uniman; password: friendintegrityfaith)

## 2. Preliminaries

### 2.1. Multiplicative Preference Relation

In this section, we review the form of multiplicative preference relation (MPR) (Satty, 1980) and its consistency definitions. For a set of alternatives $X = \{x_1, x_2, ..., x_n\}$, the definition of MPR can be presented as:

**Definition 2.1** (Satty, 1980). A MPR $R$ on $X$ is expressed by $R = (r_{ij})_{n \times n} \subset X \times X$, where $r_{ij}$ is denoted by Saaty's 1-9 scale indicating the preference degree of $x_i$ over $x_j$, and satisfies $r_{ij} r_{ji} = 1$ for $i, j = 1, 2, ..., n$.

$r_{ij} = 1$ represents indifference between $x_i$ and $x_j$, $r_{ij} > 1$ represents $x_i$ is superior to $x_j$, where $r_{ij} < 1$ represents $x_i$ is inferior to $x_j$ (Satty, 1977). $R$ is perfectly consistent if (Satty, 1980)

$$r_{ij} = \frac{w_i}{w_j} \tag{1}$$

where $w = (w_1, w_2, ..., w_n)^T$ is the priority vector derived from $R$. Furtherly, Crawford and Williams (1985) proposed the GCI for $R$:

$$GCI(R) = 1 - \frac{2}{(n-1)(n-2)} \sum_{i,j=1, i<j}^{n} (\log(r_{ij}) - \log(w_i) + \log(w_j)) \tag{2}$$

### 2.2. Linguistic Preference Relation

Considering that people are familiar with language messages, Zadeh (1975) introduced the linguistic term set (LTS), and noted its basic form as $S = \{s_\alpha | \alpha = 0,1,...,2\tau\}$, where $\tau$ is a positive number and $s_\alpha$ is a linguistic term (LT). Operations of any two LTs $s_\alpha$ and $s_\beta$ are (Zadeh, 1975; Xu, 2005):

(1) Addition: $s_\alpha \oplus s_\beta = s_{\alpha+\beta}$;

(2) Scalar multiplication: $\lambda s_\alpha = s_{\lambda\alpha}$ ($\lambda \geq 0$);

(3) Negation operation: $Neg(s_\alpha) = s_{2\tau-\alpha}$.

The definition of LPR was furtherly proposed as:

**Definition 2.2** (Herrera et al., 1995). Let $X = \{x_1, x_2,..., x_n\}$ be a set of alternatives, a LPR is denoted by $L = (l_{ij})_{n \times n} \subset X \times X$, where $l_{ij} \in S$ indicating the preference degree of $x_i$ over $x_j$. $L$ is reciprocal if $l_{ij} = Neg(l_{ji})$ for $i, j = 1, 2,..., n$.

*2.3. Hesitant Fuzzy Linguistic Preference Relation*

Let $X = \{x_1, x_2,..., x_n\}$ be a set of alternatives, then a hesitant fuzzy linguistic term set (HFLTS) (Rodríguez, 2012) is an ordered finite subset of several consecutive LTs of an LTS $S = \{s_\alpha | \alpha = 0,1,...,2\tau\}$. Liao and Xu (2015) gave its mathematical notation as

$$b = \{\langle x_i, b(x_i)\rangle | x_i \in X\} \tag{3}$$

where $b(x_i)$ is called a hesitant fuzzy linguistic element.

Later on, Zhu and Xu (2014a) defined the HFLPR as follows:

**Definition 2.3** (Zhu & Xu, 2014a). Let $X = \{x_1, x_2,..., x_n\}$ be a set of alternatives, an HFLPR on $X$ is represented by $B = (b_{ij})_{n \times n} \subset X \times X$, where $b_{ij} = \{b_{ij}^\ell | \ell = 1,..., \#b_{ij}\}$ ($\#b_{ij}$ is the number of LTs in $b_{ij}$), is a HFLTS, and satisfies

$$b_{ij}^{o(\ell)} \oplus b_{ji}^{o(\ell)} = s_0, \ b_{ii} = s_0, \ \#b_{ij} = \#b_{ji}, \ b_{ij}^{o(\ell)} < b_{ij}^{o(\ell+1)}, \ b_{ji}^{o(\ell+1)} < b_{ji}^{o(\ell)}, \ \forall i,j = 1,2,...,n \tag{4}$$

where $b_{ij}$ denotes the hesitance preference degrees $x_i$ over $x_j$, and $o(\ell)$ is the order of $\ell$ th LT in $b_{ij}$.

For ease of calculation, suppose that $b^+$ and $b^-$ are respectively the maximum and minimum LTs of a HFLTS $b$, then a normalization way for $b$ was proposed to add extra LT(s) in $B$, where the added elements are obtained by $\bar{b} = \varsigma b^+ \oplus (1-\varsigma) b^-$ ($0 \leq \varsigma \leq 1$) (Zhu & Xu, 2014a). Let $b_\alpha = \{b_\alpha^\ell | \ell = 1, 2, ..., \#b_\alpha\}$ and $b_\beta = \{b_\beta^\ell | \ell = 1, 2, ..., \#b_\beta\}$ be two HFLTSs with the same length, which associate with two ordered lower index sets $I(b_\alpha)$ and $I(b_\beta)$, then (Zhu & Xu, 2014a)

(1) Addition: $I(b_\alpha) + I(b_\beta) = \bigcup_{\substack{I^{o(\ell)}(b_\alpha) \in I(b_\alpha) \\ I^{o(\ell)}(b_\beta) \in I(b_\beta)}} \{I^{o(\ell)}(b_\alpha) + I^{o(\ell)}(b_\beta)\}$;

(2) Scalar multiplication: $\lambda I(b_\alpha) = \bigcup_{I^{o(\ell)}(b_\alpha) \in I(b_\alpha)} \{\lambda I^{o(\ell)}(b_\alpha)\}$ ($\lambda \geq 0$);

(3) Power: $(I(b_\alpha))^\lambda = \bigcup_{I^{o(\ell)}(b_\alpha) \in I(b_\alpha)} \{(I^{o(\ell)}(b_\alpha))^\lambda\}$ ($\lambda \geq 0$).

## 3. Consistency and consensus driven GDM with HFLPRs

In the section, we establish the framework of GDM with HFLPRs by addressing the geometric consistency of a HFLPR, providing a consensus reaching process and deriving the outcomes.

### *3.1. Geometric consistency of a HFLPR*

Geometric consistency is a tool proposed by Crawford and Williams (1985) for supporting DMs to make a reasonable decision with a MPR, which is developed by utilizing the row geometric mean to build an optimization model. To exert its property in handling preference relation, based on the linguistic GCI (Feng et al., 2018) and the additive transitivity property of LPR (Alonso et al., 2008), we make discussions for the GCI of HFLPR.

**Definition 3.1.** Let $X = \{x_1, x_2, ..., x_n\}$ be a set of alternatives, and $B = (b_{ij})_{n \times n}$ be a normalized HFLPR on $X$, where $b_{ij} = \{b_{ij}^\ell | \ell = 1, ..., \#b\}$, then $B$ is called a perfectly consistent HFLPR if

$$\left(I^{o(\ell)}(b_{ik}) - \tau\right) + \left(I^{o(\ell)}(b_{kj}) - \tau\right) = \left(I^{o(\ell)}(b_{ij}) - \tau\right), \quad i, j, k = 1, 2, ..., n \quad (5)$$

**Theorem 3.1.** Suppose that $B = (b_{ij})_{n \times n}$ is a consistent HFLPR, where $b_{ij} = \{b_{ij}^\ell | \ell = 1, ..., \#b\}$, if we let

$$h_{ij} = \frac{I(b_{ij})}{2\tau} \quad (6)$$

then $H = (h_{ij})_{n \times n}$ is a consistent hesitant fuzzy preference relation (HFPR) defined by Xia and Xu (2013), where $h_{ij} = \{h_{ij}^{\ell} | \ell = 1,...,\#h\}$, $\#b = \#h$.

**Proof.** For a consistent HFLPR, by Definition 3.1, we get that $I^{o(\ell)}(b_{ik}) + I^{o(\ell)}(b_{kj}) - \tau = I^{o(\ell)}(b_{ij})$, then for a consistent HFPR $H = (h_{ij})_{n \times n}$, we get

$$h_{ik}^{\ell} + h_{kj}^{\ell} - 0.5 = \frac{I^{o(\ell)}(b_{ik})}{2\tau} + \frac{I^{o(\ell)}(b_{kj})}{2\tau} - \frac{\tau}{2\tau} = \frac{I^{o(\ell)}(b_{ij})}{2\tau} = h_{ij}^{\ell}.$$

Conversely, $I^{o(\ell)}(b_{ik}) + I^{o(\ell)}(b_{kj}) = 2\tau h_{ik}^{\ell} + 2\tau h_{kj}^{\ell} = 2\tau h_{ij}^{\ell} + \tau = I^{o(\ell)}(b_{ij}) + \tau$, which completes the proof of Theorem 3.1.

**Theorem 3.2.** Let $X = \{x_1, x_2,...,x_n\}$ be a set of alternatives, $B = (b_{ij})_{n \times n}$ be a HFLPR on $X$, then $B$ is a consistent HFLPR if

$$I^{o(1)}(b_{ij}) \, or \, ... \, or \, I^{o(\#b)}(b_{ij}) = 2\tau(\alpha(w_i - w_j) + 0.5) \tag{7}$$

**Proof.** By Theorem 3.1, we have

$$I^{o(\ell)}(b_{ij}) = I^{o(\ell)}(b_{ik}) + I^{o(\ell)}(b_{kj}) - \tau.$$

Based on a consistent hesitant multiplicative preference relation $M = (m_{ij})_{n \times n}$, where $m_{ij} = \{m_{ij}^{\ell} | \ell = 1,...,\#m\}$ satisfies $\frac{w_i}{w_j} = m_{ij}^{(1)} \, or \, ... \, or \, m_{ij}^{(\#m)}$, $i, j = 1, 2,..., n$ (Zhu & Xu, 2014b), and a consistent fuzzy preference relation (FPR) $A = (a_{ij})_{n \times n}$ satisfies $a_{ij} = \alpha(w_i - w_j) + 0.5$ (Chiclana et al., 1998), we deduce that

$$\alpha(w_i - w_j) + 0.5 = h_{ij}^{(1)} \, or \, ... \, or \, h_{ij}^{(\#h)}.$$

where the parameter $\alpha$ is determined by DMs for reflecting the importance to the difference between $w_i$ and $w_j$ satisfying $\alpha \geq \frac{n-1}{2}$ (Lu, 2002). Furtherly

$$2\tau(\alpha(w_i - w_j) + 0.5) = I^{o(1)}(b_{ij}) \, or \, ... \, or \, I^{o(\#b)}(b_{ij}),$$

which completes the proof of Theorem 3.2.

Denote $\varphi_{ij} = I^{o(1)}(b_{ij}) \, or \, ... \, or \, I^{o(\#b)}(b_{ij})$, then the GCI for a HFLPR can be given as follows:

**Definition 3.2.** Let $X = \{x_1, x_2, \ldots, x_n\}$ be a set of alternatives, $B = (b_{ij})_{n \times n}$ be a HFLPR on $X$, and $w = (w_1, w_2, \ldots, w_n)^T$ be the priority vector derived by the row geometric mean method, then the hesitant fuzzy linguistic geometric consistency index (HFLGCI) is defined as

$$HFLGCI = \min\left\{\frac{2}{(n-1)(n-2)}\sum_{j>i}\left[\frac{\varphi_{ij}}{\tau} - 2\alpha(w_i - w_j) - 1\right]^2\right\} \tag{8}$$

It is evident that $HFLGCI = 0$ indicates $B$ is perfectly consistent. Furthermore, suppose that the critical value of HFLGCI is $\overline{HFLGCI}$, then

(1) If the HFLGCI of $B$ is greater than $\overline{HFLGCI}$, then $B$ is unacceptably consistent;

(2) If the HFLGCI of $B$ is less than or equal to $\overline{HFLGCI}$, then $B$ is acceptably consistent.

**Proposition 3.1.** Let $B = (b_{ij})_{n \times n} \subset X \times X$ be a HFLPR, where $b_{ij} = \{b_{ij}^l | l = 1, \ldots, \#b\}$, then the available solutions, which follow as a collection of priority vectors $w^\ell = (w_1^\ell, w_2^\ell, \ldots, w_n^\ell)^T$ ($\ell = 1, \ldots, \#b$), based on the HFLGCI, where

$$w_i^l = \exp\left(\ln\left(\prod_{j=1}^n 9^{\left(\frac{I(\varphi_{ij})}{\tau}-1\right)}\right)^{\frac{1}{n}} - \ln\left(\sum_{p=1}^n \left(\prod_{j=1}^n 9^{\left(\frac{I(\varphi_{pj})}{\tau}-1\right)}\right)^{\frac{1}{n}}\right)\right) \tag{9}$$

We rely on the relationship among different types of preference relations to deduce Proposition 3.1. Based on the relationship between a MPR $R = (r_{ij})_{n \times n}$ and a FPR $A = (a_{ij})_{n \times n}$: $r_{ij} = 9^{2a_{ij}-1}$ (Xu, 2002), and the relationship between a consistent FPR $A = (a_{ij})_{n \times n}$ and a consistent LPR $L = (l_{ij})_{n \times n}$: $a_{ij} = \frac{I(l_{ij}) + \tau}{2\tau}$ (Feng et al., 2018), we can get $r_{ij} = 9^{\left(\frac{I(l_{ij})}{\tau}\right)}$. For a consistent HFLPR $B = (b_{ij})_{n \times n}$, where $b_{ij} = \{b_{ij}^\ell | \ell = 1, \ldots, \#b\}$, according to Definition 3.1, we know that $B$ contains $\#b$ perfect LPRs. Since the optimal solution of a MPR $R$ is $w_i = \frac{\sqrt[n]{\prod_{j=1}^n r_{ij}}}{\sum_{p=1}^n \sqrt[n]{\prod_{j=1}^n r_{pj}}}$ and $r_{ij} = 9^{\left(\frac{I(l_{ij})}{\tau}\right)}$, Equation (10) can be obtained.

**Proposition 3.2.** Let $B = (b_{ij})_{n \times n} \subset X \times X$ be a HFLPR, then based on Proposition 3.1, the optimal solutions based on the HFLGCI can be extracted as $w^* = (w_1^*, w_2^*, ..., w_n^*)^T$, which corresponds to the priority vector that makes the HFLGCI minimum.

**Proposition 3.3.** Let $B = (b_{ij})_{n \times n} \subset X \times X$ be a HFLPR, then a perfectly consistent LPR can be derived from $B$, denoted as $\overline{L} = (\overline{l}_{ij})_{n \times n}$, where

$$\overline{l}_{ij} = I^{-1}\left(2\tau\left(\frac{w_i^*}{w_i^* + w_j^*}\right)\right) \qquad (10)$$

where $w_i^*$ and $w_j^*$ for $i, j = 1, 2, ..., n$ are the priorities derived from HFLPR that corresponds to the minimum value of HFLGCI.

The procedure of consistency checking and inconsistency improving for a HFLPR $B = (b_{ij})_{n \times n} \subset X \times X$ can be summarized as:

**Algorithm I.** An algorithm for obtaining an acceptable HFLPR

Input: The HFLPR $B = (b_{ij})_{n \times n}$, where $b_{ij} = \{b_{ij}^\ell | \ell = 1, ..., \#b\}$, the critical value $\overline{HFLGCI}$.

Output: The HFLPR $B$ or the revised $B$ with acceptable consistency.

*Step 1.* Let $t = 1$, denote $B = (b_{ij})_{n \times n}$ as $B(t) = (b_{ij}(t))_{n \times n}$;

*Step 2.* Calculate the priority weight vectors of $B(t)$ by Equation (9);

*Step 3.* Compute *HFLGCI* of $B(t)$ by Equation (8) for each priority weight vector. The priority vector that makes the *HFLGCI* minimum is the optimal solution, let the minimum *HFLGCI* be the consistency index of $B(t)$.

*Step 4.* If $HFLGCI > \overline{HFLGCI}$, then go the next step; otherwise, go to Step 6;

*Step 5.* Obtain perfectly consistent LPR $\overline{L} = (\overline{l}_{ij})_{n \times n}$ from $B(t)$ by Equation (10). Based on $\overline{L}$, adjust each $b_{ij}(t)$ in $B(t)$ to derive a new HFLPR by using $I(b_{ij}^{o(\ell)}(t+1)) = \beta I(b_{ij}^{o(\ell)}(t)) + (1-\beta)I(\overline{l}_{ij})$ ($0 \leq \beta \leq 1$), denoted as $B(t+1)$. Let $t = t+1$. Return to Step 2;

*Step 6*. Output $B(t)$ and $t$.

End.

**Proposition 3.4.** The convergence and monotonicity hold in Algorithm I.

**Proof.** (1) Monotonicity:

In Step 5, the LPR derived from the modified HFLPR with the minimum consistency deviation is closer to the correspondingly perfect LPR. Based on Equation (5) and Equation (8), we get

$$HFLGCI(t) \leq HFLGCI(t-1).$$

It indicates that the value of consistency degree holds or decreases after each loop of step 5, which proves the Monotonicity of Algorithm I.

(2) Convergence:

We assume that $\Delta$ is the minimum value change consistency degree during the $t$ th loop of Step 5, then the following equation holds

$$HFLGCI(t) \leq HFLGCI(t-1) - \Delta,$$

where $t$ is the looping time of Step 5. Assuming $HFLGCI(0)$ is the consistency degree of the original HFLPR, then

$$HFLGCI(t) \leq HFLGCI(t-1) - \Delta \leq HFLGCI(t-2) - 2\Delta \leq ... \leq HFLGCI(0) - t\Delta.$$

In the following, we use proof by contradiction to present the convergence of Algorithm I.

If Algorithm I loops perpetually, i.e., $t \to +\infty$, then $HFLGCI(t) \leq HFLGCI(0) - t\Delta \leq -\infty$, which is contradictory with $HFLGCI(t) \geq 0$. Therefore, we can deduce that Algorithm I stops within finite steps, indicating the convergence of Algorithm I holds.

The above work completes the proof of Proposition 3.4.

The time complexity of Algorithm I is $O\left(\log_{\left(\frac{1}{0.6}\right)} 9 \times \left(n^3 + 2n^2\right)\right) = O\left(n^3\right)$ where $n$ is the number of alternatives. The maximum units of memory space required to run Algorithm I is $3 \times 9 \times n^2 + 9 \times n + 10$, where $3 \times 9 \times n^2$ units of space is for $B(t)$, $\bar{L}$ and $B(t+1)$, $9 \times n$ units of space is for the priority weight vectors, and 10 units of space is for the *HFLGCI*s. Therefore, the space complexity of Algorithm I is $O\left(n^2\right)$. The detailed analysis of time/space complexity of Algorithm I is shown in Appendix. These mean the execution time of the our proposed algorithm is either given by a polynomial on the size/time of the input or can be bounded by such a polynomial.

### *3.2. Consensus state of HFLPRs*

Before introducing the consensus measure for a group of HFLPRs, we present how to determine the weights of DMs, which is important for obtaining the aggregated HFLPR. Motivatied by the idea that the higher weight could be assigned to the DM who provides a more logical HFLPR (Meng et al., 2021), we let $X = \{x_1, x_2, ..., x_n\}$ be a set of alternatives, $B^h = (b_{ij}^h)_{n \times n} \subset X \times X$ ($h = 1, 2, ..., k$) be a series of HFLPRs on $X$ assigned by $k$ DMs, then the weight of a DM can be computed by

$$p_h = \frac{sim(B^h, \tilde{B}^h)}{\sum_{h=1}^{k} sim(B^h, \tilde{B}^h)}, \forall h = 1, 2, ..., k \tag{11}$$

where $\tilde{B}^h = (\tilde{b}_{ij}^h)_{n \times n} \subset X \times X$ ($h = 1, 2, ..., k$) are the perfect HFLPRs corresponding to the ones giving by DMs, calculated by the meanings in (Zhu & Xu, 2014a; Zhang & Wu, 2014); $sim(B^h, \tilde{B}^h)$ is the similarity measure between $B^h$ and $\tilde{B}^h$ calculated by the meanings in (Liao.et al., 2014; Liao & Xu, 2015).

### *Part 2. Consensus measure and feedback mechanism*

Since Lehrer and Wagner (2015) indicated that a rational group consensus reaching is a procedure for legitimately inspiring DMs to change their individual preferences, rather than pooling or aggregation, the section proposes a worst consensus measure and a feedback mechanism for improving the group consensus.

**Definition 3.3.** Let $X = \{x_1, x_2, ..., x_n\}$ be a set of alternatives, $B^h = (b_{ij}^h)_{n \times n} \subset X \times X$ ($h = 1, 2, ..., k$) be a series of HFLPRs on $X$ assigned by $k$ DMs. Suppose that $\bar{B}_{ij} = (\bar{b}_{ij})_{n \times n} \subset X \times X$ is the collective HFLPR aggregated by all perfect HFLPRs of $B^h$ for $h = 1, 2, ..., k$, then the worst consensus degree of the group can be measured by

$$wcd = \min_h \{sim(B^h, \bar{B})\}, \forall h = 1, 2, ..., k \tag{12}$$

**Definition 3.4.** Let $\gamma$ be the consensus threshold assigned by DMs in a GDM problem, and $B^h = (b_{ij}^h)_{n \times n} \subset X \times X$ ($h = 1, 2, ..., k$) be a series of HFLPRs on $X$, then the group is called to be consensus achieved if

$$wcd \geq \gamma, \quad \forall h = 1, 2, ..., k \tag{13}$$

Note Wu and Xu (2016) pointed out that there is no unified rule for the determination of $\gamma$. It should be confirmed under the specific decision-making environment. It is apparent that in a particular problem, the condition of group consensus may be unsatisfied with the original HFLPRs of DMs, i.e., $wcd < \gamma$, then a procedure is necessary to improve consensus. A consensus improvement mechanism usually includes computing group consensus index, judging the consensus degree, returning information, and modifying judgments. Based on these, we present a consensus improvement mechanism for a group of HFLPRs as follows:

*Part 1. Group consensus index*

Before measuring the group consensus, firstly, the collective HFLPR should be calculated. Suppose that $\tilde{B}^h = (\tilde{b}_{ij}^h)_{n \times n}$ are the perfectly consistent ones of $B^h = (b_{ij}^h)_{n \times n}$ for $h = 1, 2, ..., k$, then the collective HFLPR $\bar{B}_{ij} = (\bar{b}_{ij})_{n \times n}$ can be obtained by the *HFLWA* operator (Xu et al., 2015), where

$$\bar{b}_{ij} = \bigoplus_{h=1}^{k} p_h \tilde{b}_{ij}^h = \bigcup_{(\tilde{b}_{ij}^h)^{o(\ell)} \in \tilde{b}_{ij}^h} \left\{ \sum_{h=1}^{k} p_h (\tilde{b}_{ij}^h)^{o(\ell)} \right\} \tag{14}$$

The worst consensus degree $wcd$ can be calculated by Equation (12) and Equation (14).

*Part 2. Consensus judgment*

Let $\gamma$ be the critical value of acceptable consensus assigned by DMs, then

(1) if $wcd \geq \gamma$, then the group consensus is achieved;

(2) if $wcd < \gamma$, then the group consensus is unsatisfied, which should be repaired.

*Part 3. Judgment modification*

Different from other modification processes, the modification here focuses on addressing the individual HFLPR on an alternative with the maximal distance from the collective HFLPR. More specifically,

(1) Find the maximal distance of HFLPR on alternative from the collective HFLPR by

$$md = \max_{h}\{\sum_{j=1}^{n} dis(b_{ij}^{h}, \bar{b}_{ij})\} \tag{15}$$

where $dis(b_{ij}^{h}, \bar{b}_{ij})$ is the distance measure between $b_{ij}^{h}$ and $\bar{b}_{ij}$ computed by the means in (Liao et al., 2014).

(2) Modify the HFLPR on the alternative selected by Equation (15) as

$$(b_{ij}^{h})_{New} = \zeta b_{ij}^{h} \oplus (1-\zeta)\tilde{b}_{ij}, \ 0 \leq \zeta \leq 1 \tag{16}$$

(3) Let $b_{ij}^{h}$ be $(b_{ij}^{h})_{New}$, compute its $wcd$. If $wcd < \gamma$, then repeat to find the maximal distance on alternative and modify the corresponding judgment; otherwise, the group consensus is achieved, rank alternatives based on the collective HFLPR.

### 3.3. A procedure for consistency and consensus driven GDM with HFLPRs

This section briefly concludes the GDM process with HFLPRs based on the acceptable consistency and consensus measurements.

**Algorithm II.** An algorithm for GDM with HFLPRs based on consistency and consensus measurements.

Input: The individual HFLPRs $B^h = (b_{ij}^h)_{n \times n} \subset X \times X$ ( $h = 1, 2, ..., k$ ) given by DMs, where $b_{ij}^h = \{(b_{ij}^h)^{o(\ell)} | \ell = 1, 2, ..., \#b\}$, the critical value $\overline{HFLGCI}$, the critical value $\gamma$.

Output: A ranking on $X$.

*Step 1*. Utilize Algorithm I to obtain the acceptably consistent HFLPRs $\widehat{B}^h = (\widehat{b}_{ij}^h)_{n \times n}$ of all individual HFLPRs $B^h = (b_{ij}^h)_{n \times n}$ for $h = 1, 2, ..., k$;

*Step 2*. Obtain the perfectly consistent HFLPRs $\tilde{B}^h = (\tilde{b}_{ij}^h)_{n \times n}$ of all individual HFLPRs $B^h = (b_{ij}^h)_{n \times n}$ for $h = 1, 2, ..., k$, compute the weight of each DM by Equation (11);

*Step 3*. Aggregate all $\tilde{B}^h$ and obtain the collective perfectly consistent HFLPR $\bar{B}_{ij} = (\bar{b}_{ij})_{n \times n}$ by Equation (14). Let $\tau = 1$, and $\widehat{B}^h = (\widehat{b}_{ij}^h)_{n \times n}$ be $\widehat{B}^h(\tau) = (\widehat{b}_{ij}^h(\tau))_{n \times n}$;

*Step 4*. For $h = 1, 2, ..., k$, calculate the worst consensus degree $wcd(\tau)$ of $\widehat{B}^h(\tau)$ ( $h = 1, 2, ..., k$ ) by Equation (12). If $wcd(\tau) < \gamma$, then go the next step; otherwise, go to Step 6;

*Step 5*. For the DM who has the worst consensus degree, Using Equation (15) and Equation (16) to modify his/her HFLPR. Let $\tau = \tau + 1$, return to Step 4;

*Step 6*. Aggregate all $\widehat{B}^h(\tau)$ ( $h = 1, 2, ..., k$ ) to obtain the collective HFLPR $\widehat{B} = (\widehat{b}_{ij})_{n \times n}$, get the optimal solution (the priority weight vector) of $\widehat{B}$ by Equation (9) and Proposition 3.2, rank alternatives according to the priority weight vector.

End.

**Proposition 3.5.** The convergence and monotonicity hold in Algorithm II.

**Proof.** This proof can be easily completed by the same idea as the proof for Proposition 3.4, which is omitted here.

The time complexity of Algorithm II is $O(3n^3k + 3n^2k + 1) = O(3n^3k)$. During the calculation of Algorithm II, $9 \times n^2k * 2$ units of memory space is used by all DM's HFLPRs and their corresponding perfectly consistent HFLPRs. The rest of memory space is required by the collective perfectly consistent

HFLPR ($9n^2$ units), the weight vector of DMs ($k$ units) and the priority weight vectors ($9n$ units), repsectively. We could conclude that the space complexity of Algorithm II is $O(n^2 k)$. The detailed analysis of time/space complexity of Algorithm II is shown in Appendix. Similar to Algorithm I, the execution time of the our proposed algorithm is either given by a polynomial on the size/time of the input or can be bounded by such a polynomial. Since our proposed algorithm is polynomial without using any NP-hard optimization method, lots of computation souce/time can be saved.

The whole procedure can be simplified in Figure 1.

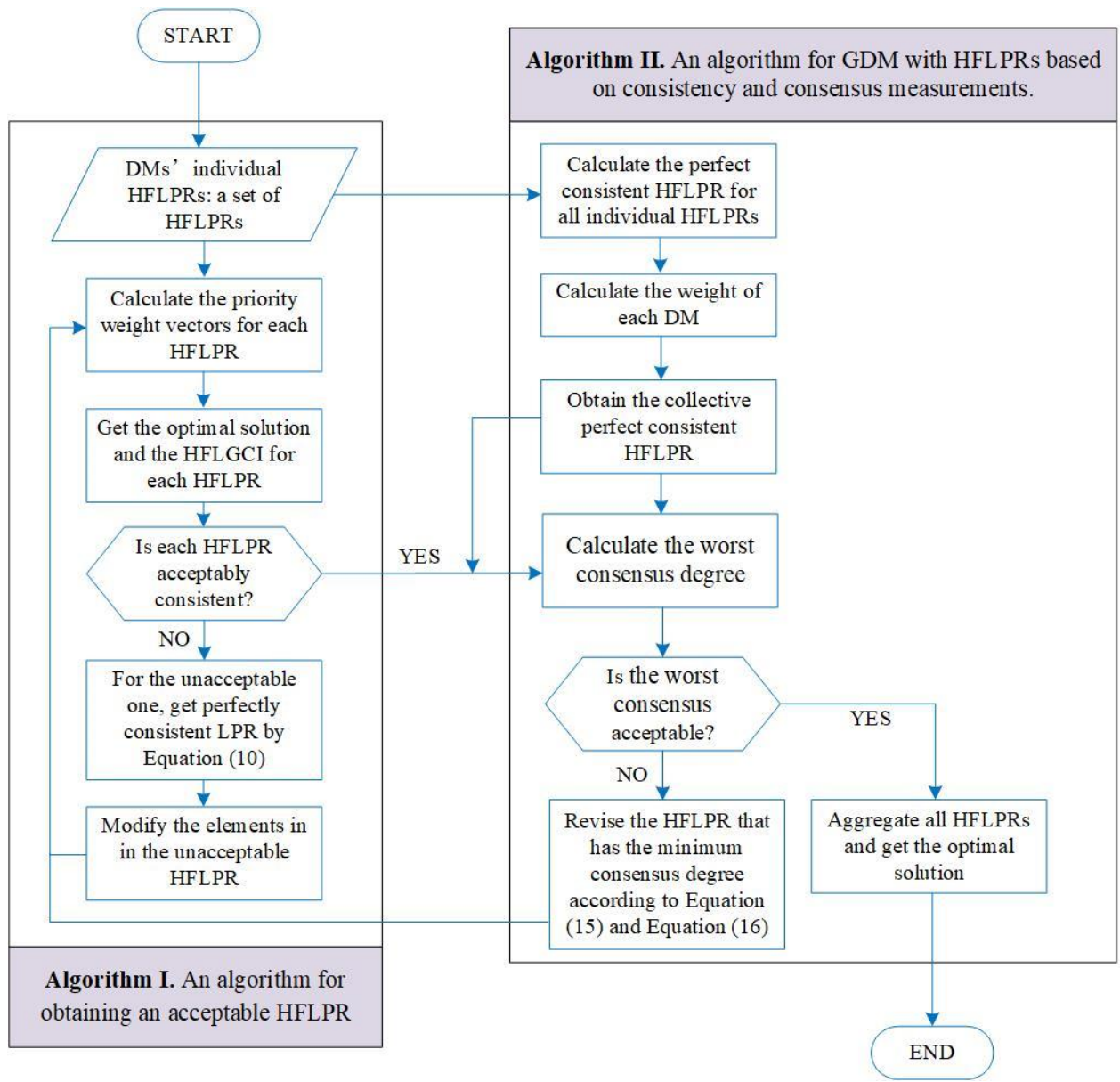

**Figure 1.** Brief diagram for consistency and consensus driven GDM with HFLPRs

## 4. Discussions

### 4.1. Estimated critical values of HFLGCI

In Algorithm I, $\overline{HFLGCI}$ is the critical value used to judge the acceptable consistency of an HFLPR, it decides that the algorithm continues to run or not. In this section, we aim to investigate the values of $\overline{HFLGCI}$ in different scenarios, which provides decision support for the proposed algorithm.

We consider $n$ from 3 to 8 to meet common decision-making situations where the number of alternatives varies from 3 to 8. Since the undetermined parameter $\alpha$ may have an impact on $\overline{HFLGCI}$ in Algorithm I, for more general, we discuss the determination of $\overline{HFLGCI}$ under different experiment scenarios, where the value of $\alpha$ is assigned as $(n-1)/2$, $(n-1)/2+0.2$, $(n-1)/2+0.4$, $(n-1)/2+0.6$, respectively. To decrease the interferences caused by specific decision-making situations, we let $\beta = 0.5$, and randomly create 1000 HFLPRs $B$ based on the LTS $S = \{s_\alpha | \alpha = 0,1,...,8\}$ for each experiment scenario (each value of $n$ and each value of $\alpha$), then run Algorithm I for each created HFLPR $B$.

To investigate the value of $\overline{HFLGCI}$ under each experiment scenario, we consider the acceptable consistency of the HFLPR from another aspect: if $t$ in Algorithm I is not equal to 1 and the difference between $HFLGCI_{t-1}$ and $HFLGCI_t$ is smaller than or equal to 0.0001, then we hold on the opinion that the HFLPR with $HFLGCI_{t-1}$ meets the requirement of acceptable consistency.

The outcomes of the above experiments with different values of $n$ and different values of $\alpha$ are presented in Table I.

**Table I.** The mean and variance of $HFLGCI_{t-1}$ derived from Algorithm I for each scenario

| $HFLGCI_{t-1}$ | $\alpha = (n-1)/2$ | | $\alpha = (n-1)/2+0.2$ | | $\alpha = (n-1)/2+0.4$ | | $\alpha = (n-1)/2+0.6$ | |
|---|---|---|---|---|---|---|---|---|
| | Mean | Variance | Mean | Variance | Mean | Variance | Mean | Variance |
| $n = 3$ | 0.0431 | 0.0021 | 0.1047 | 0.0086 | 0.1996 | 0.0246 | 0.3321 | 0.0669 |
| $n = 4$ | 0.0528 | 0.0015 | 0.0905 | 0.0036 | 0.1431 | 0.0088 | 0.2199 | 0.0181 |
| $n=5$ | 0.0595 | 0.0012 | 0.0853 | 0.0028 | 0.1234 | 0.0056 | 0.1638 | 0.0098 |
| $n=6$ | 0.0594 | 0.0011 | 0.0837 | 0.0022 | 0.1100 | 0.0036 | 0.1411 | 0.0061 |
| $n=7$ | 0.0616 | 0.0011 | 0.0771 | 0.0018 | 0.1011 | 0.0028 | 0.1217 | 0.0043 |
| $n=8$ | 0.0597 | 0.0010 | 0.0742 | 0.0015 | 0.0912 | 0.0023 | 0.1113 | 0.0032 |

Table I shows the outcomes of mean and variance of $HFLGCI_{t-1}$ for 1000 times running of Algorithm I under each value of $n$ and each value of $\alpha$, which verifies that Algorithm I is feasible and has convergence in general cases. We get from Table I that with the increase of $\alpha$, the mean of $HFLGCI_{t-1}$ increases, which correctly reflects the property presented in (Lu, 2002), i.e., the smaller value of $\alpha$ implies that DM cares more about the difference among alternatives (the priority difference of two alternatives is larger).

Since the determination of $\alpha$ is related to the value of $n$, we cannot directly compare each column in Table I. According to (Lu, 2002), when the number of alternatives in HFLPR increases, the differences among alternatives decrease. Reflect in Equation (8), the value of $w_i - w_j$ gets smaller when $n$ increases, which means when $\alpha$ increases, the value of HFLGCI becomes less sensitive with the increase of $n$, which leads $HFLGCI_{t-1}$ being less sensitive with the increase of $n$. To intuitively reflect this situation, we let the second, the third, and the fourth columns of Table I subtract their previous columns and obtain Table II as follows:

**Table II.** The change of $HFLGCI_{t-1}$ when $\alpha$ ascends

| The change of $HFLGCI_{t-1}$ | $\alpha$ increases 0.2 from $(n-1)/2$ | | $\alpha$ increases 0.2 from $(n-1)/2+0.2$ | | $\alpha$ increases 0.2 from $(n-1)/2+0.4$ | |
|---|---|---|---|---|---|---|
| | Mean | Variance | Mean | Variance | Mean | Variance |
| $n=3$ | 0.0616 | 0.0065 | 0.0949 | 0.0160 | 0.1325 | 0.0423 |
| $n=4$ | 0.0377 | 0.0021 | 0.0526 | 0.0052 | 0.0768 | 0.0093 |
| $n=5$ | 0.0258 | 0.0016 | 0.0381 | 0.0028 | 0.0404 | 0.0042 |
| $n=6$ | 0.0243 | 0.0011 | 0.0263 | 0.0014 | 0.0311 | 0.0033 |
| $n=7$ | 0.0155 | 0.0007 | 0.0240 | 0.0010 | 0.0206 | 0.0015 |
| $n=8$ | 0.0145 | 0.0005 | 0.017 | 0.0008 | 0.0201 | 0.0009 |

By Table II, we could find that the changes of mean and variance of $HFLGCI_{t-1}$ becoming smaller when $n$ increases in each column. On the contrary, from each row of Table II, the change of $HFLGCI_{t-1}$ gets more sensitive with the increase of $\alpha$, indicating that the importance of the priority difference $w_i - w_j$ in Equation (8) decreases when $\alpha$ increases. Therefore, we recommend the DM to set the value of $\alpha$ as $(n-1)/2$ in reality, which is the most effective way to emphasize the importance of the priority differences among all alternatives. To better reflect the trend of data changes in Table II, we draw Figure 2. To further discuss the suggestive value of $\overline{HFLGCI}$ according to the above-collected values of $HFLGCI_{t-1}$, we show the distribution density of $HFLGCI_{t-1}$ in Figure 3.

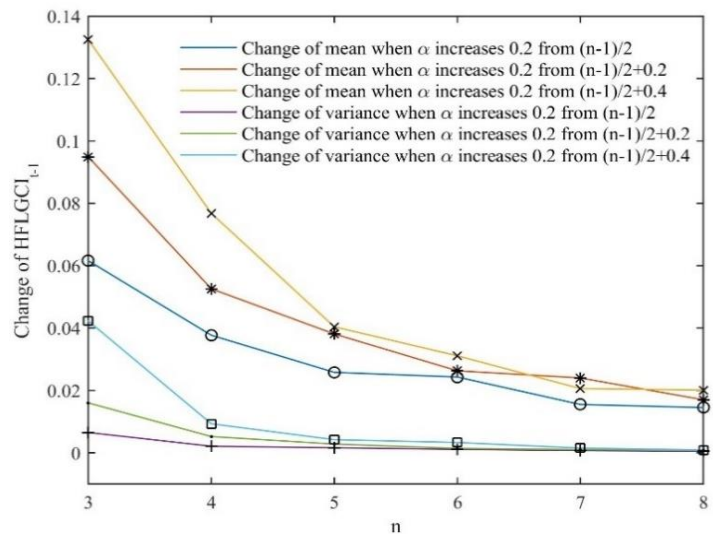

**Figure 2.** The change of $HFLGCI_{t-1}$ when $\alpha$ ascends

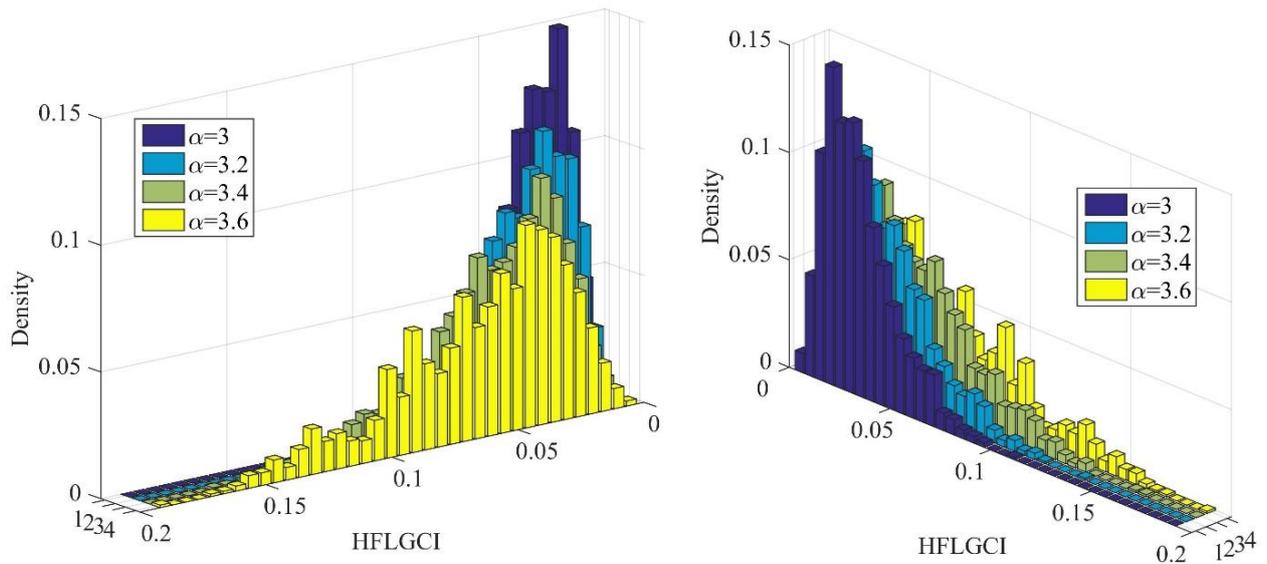

**Figure 3.** The distribution density of $HFLGCI_{t-1}$ when $n=7$ (bar chart)

Figure 3 manifests the experiment scenarios that the distribution density of $HFLGCI_{t-1}$ with different values of $\alpha$ when $n=7$. It is noted that the distribution densities of $HFLGCI_{t-1}$ with different values of $\alpha$ and $n$ present similar graphical shapes. The distribution of the experiment scenario with a smaller value of $\alpha$ is more narrow and closer to the $y$ axis compared to other experiment scenarios in Figure 3. For better understanding Figure 3, we use the polynomial curve fitting to create a distribution density curve for each $\alpha$ and redraw Figure 4.

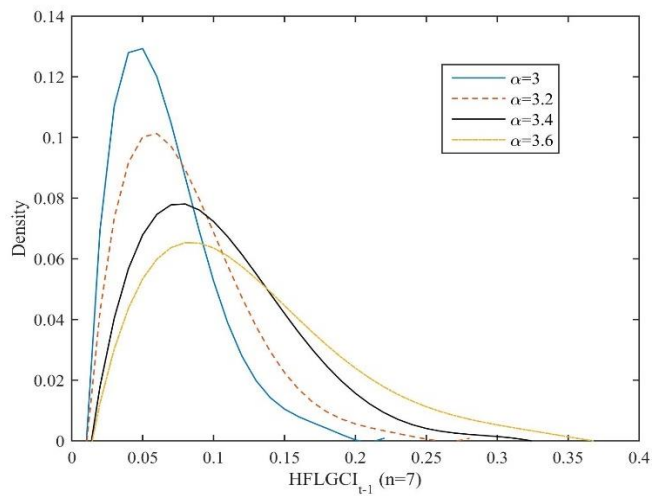

**Figure 4.** Distribution density of $HFLGCI_{t-1}$ when $n=7$ (line chart)

By Figure 4, we could see that these four experiment scenarios' peak value moving to the left with the increase of $\alpha$. It should be noted that the right side of each curve is similar to the normal distribution curve in Figure 4, even though the whole distribution curves are not entirely similar to the normal distribution curve. The reason is that the HFLGCI is always a positive number, and the setted stop condition as the difference between $HFLGCI_t$ and $HFLGCI_{t-1}$ is not greater than 0.0001. In that condition, some outcomes of $HFLGCI_{t-1}$ which located on the right side of the mean values may not be mathematically optimal (they may have smaller values in our random experiment); some outcomes of $HFLGCI_{t-1}$ are only convergent to zero and do not manifest as a negative number. With such analyses and according to the central limit theorem, we can still treat the right side of all distribution curves as the normal distribution curves. Therefore, we can conclude the suggested $\overline{HFLGCI}$ of each scenario in Table I as $mean + 3\sqrt{variance}$, which could cover 99.73% of cases and make sure Algorithm I could stop and be convergent to a reasonable result. The suggestive values of $\overline{HFLGCI}$ are listed in Table III.

**Table III.** The suggestive values of $\overline{HFLGCI}$

| $\overline{HFLGCI}$ | $\alpha = \dfrac{n-1}{2}$ | $\alpha = \dfrac{n-1}{2} + 0.2$ | $\alpha = \dfrac{n-1}{2} + 0.4$ | $\alpha = \dfrac{n-1}{2} + 0.6$ |
|---|---|---|---|---|
| $n = 3$ | 0.1816 | 0.3836 | 0.6704 | 1.1081 |
| $n = 4$ | 0.1559 | 0.2708 | 0.4248 | 0.6230 |
| $n = 5$ | 0.1738 | 0.2448 | 0.3477 | 0.4604 |
| $n = 6$ | 0.1606 | 0.2228 | 0.2908 | 0.3748 |
| $n = 7$ | 0.1625 | 0.2051 | 0.2586 | 0.3182 |
| $n = 8$ | 0.1537 | 0.1914 | 0.2360 | 0.2799 |

*4.2. Comparative analyses*

In this section, we make comparative analyses among the proposed Algorithm II and three existing similar works. Firstly, we make general comparative analyses in Table IV.

Table IV. Comparative analysis of four GDM methods

|  | Method in (Wu et al., 2019b) | Method in (Wu & Xu, 2016) | Method in (Zhang & Chen, 2019) | Proposed Algorithm II |
|---|---|---|---|---|
| Considering all possible LPRs | Yes | No | Yes | Yes |
| Disregarding some values | No | Yes | No | No |
| Adding extra values | No | No | No | No |
| Checking consistency | Yes | Yes | Yes | Yes |
| Improving consistency | Yes | Yes | Yes | Yes |
| Checking Consensus | Yes | Yes | Yes | Yes |
| Improving consensus | Yes | Yes | Yes | Yes |
| Determining DMs' weights | Yes | No | Yes | Yes |
| The selection process | Yes | Yes | Yes | Yes |
| Calculating the priority weights from consistency analysis | Yes | Yes | Yes | Yes |
| Consistency property | Additive | Additive | Multiplicative | Additive |
| Selection basis | HFL averaging operator | HFL averaging operator | Priority weight | HFL geometric operator and priority weight |

Secondly, we apply Algorithm II to handle Example 8 in (Wu et al., 2019b), the example of Section 5 in (Wu & Xu, 2016), and Example 5.1 in (Zhang & Chen, 2019), respectively, then the outcomes can be presented in Table V. From the perspectives of iteration numbers to reach consistency and consensus, we give the comparisons between each above existing work and the proposed Algorithm II in Table VI and Table VII.

Table V. The outcomes of examples obtained by different methods

| Outcomes | Weight of each DM | Priority of each alternative | Ranking orders |
|---|---|---|---|
| Method in (Wu et al., 2019b) | 0.2526, 0.2640, 0.2464, 0.2370 | 0.0313, 0.0477, 0.0361, 0.0571 | $X_1 \succ X_3 \succ X_2 \succ X_4$ |
| The proposed Algorithm II | 0.2531, 0.2520, 0.2440, 0.2409 | 0.3636, 0.1840, 0.2786, 0.1738 | $X_1 \succ X_3 \succ X_2 \succ X_4$ |
| Method in (Wu & Xu, 2016) | 0.25, 0.25, 0.25, 0.25 | 0.0313, 0.0180, 0.0180, 0.1052 | $X_1 \succ X_3 \succ X_2 \succ X_4$ |
| The proposed Algorithm II | 0.2565, 0.2417, 0.2575, 0.2443 | 0.3342, 0.1814, 0.2851, 0.1893 | $X_1 \succ X_3 \succ X_4 \succ X_2$ |
| | | | |
| Method in (Zhang & Chen, 2019) | 0.2569, 0.3716, 0.3715 | 0.3226, 0.2554, 0.2122, 0.2099 | $X_1 \succ X_2 \succ X_3 \succ X_4$ |
| The proposed Algorithm II | 0.3285, 0.3397, 0.3319 | 0.3116, 0.3030, 0.1609, 0.2245 | $X_1 \succ X_2 \succ X_4 \succ X_3$ |

**Table VI.** The efficiency comparison based on the iteration numbers to reach consistency

| Outcomes | Critical value of $\overline{HFLGCI}$ | Number of Iterations to adjust HFLPR with acceptable consistency |
|---|---|---|
| Method in (Zhang & Chen, 2019) | 0.1 | 1, 7 and 4 rounds for 3 HFLPRs respectively |
| The proposed Algorithm II | 0.1 | 1, 2 and 2 rounds for 3 HFLPRs respectively |

**Table VII.** The efficiency comparison of different methods based on the iteration numbers to reach consensus

| Outcomes | Critical value of acceptable consensus | Number of iterations to reach consensus |
|---|---|---|
| Method in (Wu et al., 2019b) | 0.8 | 3 |
| The proposed Algorithm II | 0.85 (0.90) | 1 (3) |
| | | |
| Method in (Wu & Xu, 2016) | 0.8 | 3 |
| The proposed Algorithm II | 0.85 (0.90) | 1 (3) |

By Table IV-VII we can get the following results:

(1) The optimal alternatives obtained by the proposed algorithm and the method in (Wu & Xu, 2016) are the same. Due to the priorities of $X_2$ and $X_4$ obtained by the two methods are quite close, the obtained rankings are slightly different.

(2) The rankings derived by the proposed algorithm and the method in (Wu et al., 2019b) present the same one, although the two priority vectors of alternatives are slightly different.

(3) The rankings derived by the proposed algorithm and the method in (Zhang & Chen, 2019) are the same except the order between $X_3$ and $X_4$. Zhang and Chen's method obtains quite close priorities for alternatives $X_3$ and $X_4$. Due to the different methods for DMs' weights determination, the values in terms of weights of DMs are quite different by the two methods in Table IV.

(4) Set the critical value of acceptable consistency as 0.1, then for Example 5.1 in (Zhang & Chen, 2019), after adjusting the individual HFLPR $B^1$ for one time, the individual HFLPR $B^2$ for seven times, and the individual HFLPR $B^3$ for four times, the HFLPRs with acceptable consistency can be obtained; where Algorithm 2 only need 1, 2 and 2 times to adjust the individual HFLPRs into acceptable ones. This shows that Algorithm 2 is more efficient than the method in (Zhang & Chen, 2019).

(5) Even though both of critical values of acceptable consensus in (Wu et al., 2019b) and (Wu & Xu, 2016) are set to 0.8, which is smaller than the critical value in Algorithm 2: 0.85, the iteration numbers of the two methods to reach consensus are 3, bigger than 1 in Algorithm 2. Furthermore, Algorithm 2 only needs 3 iterations to reach a consensus with the critical value of 0.9. These indicate that Algorithm 2 is more efficient than the methods in (Wu et al., 2019b) and (Wu & Xu, 2016).

Moreover, we can give explanations and deduce comparative conclusions for the above results as follows:

(1) Since the proposed algorithm and the method in (Wu et al., 2019b) both compute the relative weight of a DM to all DMs, they get quite close weights of DMs.

(2) The main difference between the proposed algorithm and the method in (Wu & Xu, 2016) is that the proposed algorithm calculates each DM's weight, where Wu and Xu's method just assigned the same weight to DMs when aggregating the individual HFLPRs. Since different DMs have different knowledge and experience, different importance should be assigned to each DM, and neglecting the difference may be irrational.

(3) The proposed algorithm and the method in (Zhang & Chen, 2019) have similar aspects, such as checking and improving consistency for individual HFLPRs, determining DMs' weights, and calculating the priorities for alternatives based on consistency analysis. In the proposed algorithm, we consider the similarity between an HFLPR and its perfectly consistent HFLPR when calculating DMs' weights, and all HFLPRs' consistency index is optimized enough. Comparatively, the method in (Zhang & Chen, 2019) considers the distance (similarity) of an individual HFLPR with other individual HFLPRs and then gets a confidence degree based on the distance measure, which ignores the character that the logical judgments can be used as essential references for others.

(4) Another difference between the proposed algorithm and the method in (Zhang & Chen, 2019) is that the proposed algorithm uses the additive consistency concepts for HFLPRs, while Zhang and Chen's method uses the multiplicative consistency and illustrates the additive consistency existing a limitation: the sum of two LTs might fall out of the predefined LTS. However, the proposed algorithm can avoid this limitation with the definitions and method shown in Section 3.

5. **A case study – Performance evaluation of venture capital guiding funds**

With the promotion of the economy and society, the cultivation of a strategic emerging industry has become an important target to sustain social development in China. Venture capital guiding fund, which focuses on investing in the strategic emerging industry to accelerate the concentration of high-quality venture capital, projects, technologies, and talents, can exert the effect of leverage amplification for the financial fund. In short, a venture capital guiding fund provides an efficient way for the government to conduct economic and social regulation as the market participants. In recent years, the government investment scale and the driven investment capital scale on venture capital guiding funds are thriving, which are vital for economic restructuring in China. Under such a situation, evaluating the performance of venture capital guiding funds is necessary to ensure its operational standardization and promote its prosperous development.

The first step of the evaluation issue is to determine the evaluated index system. According to the indexes discussed in (Cumming, 2005, 2007), they can be selected as:

(1) Policy efficiency, which contains the aspects of leverage effect, policy implementation, among others;

(2) Economic efficiency, which includes venture capital returns, investment situation, among others;

(3) Management efficiency, which consists of management standardization, professional management techniques.

Since the method established in Section 3 have the advantages that (1) the information needed from DMs conforms to their expression habit and (2) the algorithm can ensure the logicality and rationality of the outcomes, in the following, we apply the proposed algorithm to deal with the performance evaluation issue. Suppose an enterprise with three funds, denoted as $\{A_i | i=1,2,3\}$, and a top management team containing four experts ($\{D_k | k=1,2,3,4\}$) needs to evaluate the three funds. Firstly, experts assign the priority vector of policy efficiency, economic efficiency, and management efficiency as $\varpi = (0.3, 0.5, 0.2)^T$, then according to the performance of the three funds, experts provide their pairwise comparisons on alternatives with respect to each index as follows:

(1) For the index of policy efficiency:

$$B^{1,index1} = \begin{matrix} & A_1 & A_2 & A_3 \\ A_1 & \{s_4\} & \{s_3,s_4\} & \{s_4,s_5,s_6\} \\ A_2 & \{s_4,s_5\} & \{s_4\} & \{s_5,s_6\} \\ A_3 & \{s_2,s_3,s_4\} & \{s_2,s_3\} & \{s_4\} \end{matrix},$$

$$B^{2,index1} = \begin{array}{c} \\ A_1 \\ A_2 \\ A_3 \end{array} \begin{array}{c} A_1 \quad\quad A_2 \quad\quad A_3 \\ \left[ \begin{array}{ccc} \{s_4\} & \{s_4,s_5\} & \{s_4,s_5\} \\ \{s_3,s_4\} & \{s_4\} & \{s_4,s_5\} \\ \{s_3,s_4\} & \{s_3,s_4\} & \{s_4\} \end{array} \right] \end{array},$$

$$B^{3,index1} = \begin{array}{c} \\ A_1 \\ A_2 \\ A_3 \end{array} \begin{array}{c} A_1 \quad\quad A_2 \quad\quad A_3 \\ \left[ \begin{array}{ccc} \{s_4\} & \{s_2,s_3,s_4\} & \{s_4,s_5\} \\ \{s_4,s_5,s_6\} & \{s_4\} & \{s_6,s_7\} \\ \{s_3,s_4\} & \{s_1,s_2\} & \{s_4\} \end{array} \right] \end{array},$$

$$B^{4,index1} = \begin{array}{c} \\ A_1 \\ A_2 \\ A_3 \end{array} \begin{array}{c} A_1 \quad\quad A_2 \quad\quad A_3 \\ \left[ \begin{array}{ccc} \{s_4\} & \{s_3,s_4,s_5\} & \{s_4,s_5,s_6\} \\ \{s_3,s_4,s_5\} & \{s_4\} & \{s_4,s_5\} \\ \{s_2,s_3,s_4\} & \{s_3,s_4\} & \{s_4\} \end{array} \right] \end{array},$$

where $B^{k,index1}$ is the judgments given by expert $D_k$ with respect to policy efficiency.

(2) For the index of economic efficiency:

$$B^{1,index2} = \begin{array}{c} \\ A_1 \\ A_2 \\ A_3 \end{array} \begin{array}{c} A_1 \quad\quad A_2 \quad\quad A_3 \\ \left[ \begin{array}{ccc} \{s_4\} & \{s_5,s_6\} & \{s_4,s_5,s_6\} \\ \{s_2,s_3\} & \{s_4\} & \{s_2,s_3,s_4\} \\ \{s_2,s_3,s_4\} & \{s_4,s_5,s_6\} & \{s_4\} \end{array} \right] \end{array},$$

$$B^{2,index2} = \begin{array}{c} \\ A_1 \\ A_2 \\ A_3 \end{array} \begin{array}{c} A_1 \quad\quad A_2 \quad\quad A_3 \\ \left[ \begin{array}{ccc} \{s_4\} & \{s_4,s_5\} & \{s_5,s_6\} \\ \{s_3,s_4\} & \{s_4\} & \{s_3,s_4,s_5\} \\ \{s_2,s_3\} & \{s_3,s_4,s_5\} & \{s_4\} \end{array} \right] \end{array},$$

$$B^{3,index2} = \begin{array}{c} \\ A_1 \\ A_2 \\ A_3 \end{array} \begin{array}{c} A_1 \quad\quad A_2 \quad\quad A_3 \\ \left[ \begin{array}{ccc} \{s_4\} & \{s_4,s_5,s_6\} & \{s_4,s_5\} \\ \{s_2,s_3,s_4\} & \{s_4\} & \{s_3,s_4,s_5\} \\ \{s_3,s_4\} & \{s_3,s_4,s_5\} & \{s_4\} \end{array} \right] \end{array},$$

$$B^{4,index2} = \begin{array}{c} \\ A_1 \\ A_2 \\ A_3 \end{array} \begin{array}{c} A_1 \quad\quad A_2 \quad\quad A_3 \\ \left[ \begin{array}{ccc} \{s_4\} & \{s_5,s_6\} & \{s_5,s_6\} \\ \{s_2,s_3\} & \{s_4\} & \{s_3,s_4,s_5\} \\ \{s_2,s_3\} & \{s_3,s_4,s_5\} & \{s_4\} \end{array} \right] \end{array},$$

where $B^{k,index2}$ is the judgments given by expert $D_k$ with respect to economic efficiency.

(3) For the index of management efficiency:

$$B^{1,index3} = \begin{array}{c} \\ A_1 \\ A_2 \\ A_3 \end{array} \begin{array}{c} A_1 \quad\quad A_2 \quad\quad A_3 \end{array} \\ \left[ \begin{array}{ccc} \{s_4\} & \{s_4, s_5, s_6\} & \{s_4, s_5, s_6\} \\ \{s_2, s_3, s_4\} & \{s_4\} & \{s_4, s_5, s_6\} \\ \{s_2, s_3, s_4\} & \{s_2, s_3, s_4\} & \{s_4\} \end{array} \right],$$

$$B^{2,index3} = \begin{array}{c} \\ A_1 \\ A_2 \\ A_3 \end{array} \left[ \begin{array}{ccc} \{s_4\} & \{s_5, s_6, s_7\} & \{s_4, s_5\} \\ \{s_1, s_2, s_3\} & \{s_4\} & \{s_5\} \\ \{s_3, s_4\} & \{s_3\} & \{s_4\} \end{array} \right],$$

$$B^{3,index3} = \begin{array}{c} \\ A_1 \\ A_2 \\ A_3 \end{array} \left[ \begin{array}{ccc} \{s_4\} & \{s_5, s_6\} & \{s_5, s_6\} \\ \{s_2, s_3\} & \{s_4\} & \{s_5\} \\ \{s_2, s_3\} & \{s_3\} & \{s_4\} \end{array} \right],$$

$$B^{4,index3} = \begin{array}{c} \\ A_1 \\ A_2 \\ A_3 \end{array} \left[ \begin{array}{ccc} \{s_4\} & \{s_5, s_6, s_7\} & \{s_5, s_6\} \\ \{s_1, s_2, s_3\} & \{s_4\} & \{s_4, s_5, s_6\} \\ \{s_2, s_3\} & \{s_2, s_3, s_4\} & \{s_4\} \end{array} \right],$$

where $B^{k,index3}$ is the judgments given by expert $D_k$ with respect to management efficiency, and the HFLPRs are given based on the LTS $S = \{s_\alpha | \alpha = 0, 1, ..., 8\}$.

To solve this problem, Algorithm II needs to be applied to address the HFLPRs with respect to each index. Here we set $\alpha = 1.2$, $\beta = 0.5$ and $\gamma = 0.95$ for $n = 3$, then

**Step 1.** Input all HFLPRs $B^{h,index1} = (b_{ij}^{h,index1})_{3\times 3}$, $B^{h,index2} = (b_{ij}^{h,index2})_{3\times 3}$ and $B^{h,index3} = (b_{ij}^{h,index3})_{3\times 3}$ given by DMs, utilize Algorithm I to obtain the acceptably consistent HFLPRs $\widehat{B}^{h,index1} = (\widehat{b}_{ij}^{h,index1})_{3\times 3}$, $\widehat{B}^{h,index2} = (\widehat{b}_{ij}^{h,index2})_{3\times 3}$ and $\widehat{B}^{h,index3} = (\widehat{b}_{ij}^{h,index3})_{3\times 3}$ of all individual HFLPRs for $h = 1, 2, 3, 4$. Thereinto $\widehat{B}^{h,index2} = (\widehat{b}_{ij}^{h,index2})_{3\times 3}$ are listed as follows:

$$\widehat{B}^{1,index2} = \begin{array}{c} \\ A_1 \\ A_2 \\ A_3 \end{array} \left[ \begin{array}{ccc} \{s_4\} & \{s_{5.5408}, s_{5.6658}\} & \{s_{4.0436}, s_{4.1686}, s_{4.2936}\} \\ \{s_{2.4592}, s_{2.3342}\} & \{s_4\} & \{s_{2.3707}, s_{2.4957}, s_{2.6207}\} \\ \{s_{3.9564}, s_{3.8314}, s_{3.7064}\} & \{s_{5.6293}, s_{5.5043}, s_{5.3793}\} & \{s_4\} \end{array} \right]$$

$$\widehat{B}^{2,index2} = \begin{array}{c} \\ A_1 \\ A_2 \\ A_3 \end{array} \begin{array}{c} A_1 \\ \left[ \begin{array}{ccc} \{s_4\} & \{s_{4.6739}, s_{4.7989}\} & \{s_{4.4651}, s_{4.5901}\} \\ \{s_{3.3261}, s_{3.2011}\} & \{s_4\} & \{s_{3.5349}, s_{3.6599}, s_{3.7849}\} \\ \{s_{3.5349}, s_{3.4099}\} & \{s_{4.4651}, s_{4.3401}, s_{4.2151}\} & \{s_4\} \end{array} \right] \end{array}$$

$$\widehat{B}^{3,index2} = \begin{array}{c} \\ A_1 \\ A_2 \\ A_3 \end{array} \begin{array}{c} A_1 \\ \left[ \begin{array}{ccc} \{s_4\} & \{s_{4.4614}, s_{4.7114}, s_{4.9614}\} & \{s_{4.0832}, s_{4.3332}\} \\ \{s_{3.5386}, s_{3.2886}, s_{3.0386}\} & \{s_4\} & \{s_{3.3701}, s_{3.6201}, s_{3.8701}\} \\ \{s_{3.9168}, s_{3.6668}\} & \{s_{4.6299}, s_{4.3799}, s_{4.1299}\} & \{s_4\} \end{array} \right] \end{array}$$

$$\widehat{B}^{4,index2} = \begin{array}{c} \\ A_1 \\ A_2 \\ A_3 \end{array} \begin{array}{c} A_1 \\ \left[ \begin{array}{ccc} \{s_4\} & \{s_{5.2013}, s_{5.7013}\} & \{s_{4.8622}, s_{5.3622}\} \\ \{s_{2.7987}, s_{2.2987}\} & \{s_4\} & \{s_{3.1378}, s_{3.6378}, s_{4.1378}\} \\ \{s_{3.1378}, s_{2.6378}\} & \{s_{4.8622}, s_{4.3622}, s_{3.8622}\} & \{s_4\} \end{array} \right] \end{array}.$$

**Step 2.** Obtain the perfectly consistent HFLPRs of individual HFLPRs with respect to each index, and compute the weight vector of experts as $p = (0.2523, 0.2478, 0.2488, 0.2512)^T$.

**Step 3.** Calculate the collective perfect HFLPR, where the collective perfect HFLPR with respect to index 2 is shown below:

$$\tilde{B}^{index2} = \begin{array}{c} \\ A_1 \\ A_2 \\ A_3 \end{array} \begin{array}{c} A_1 \\ \left[ \begin{array}{ccc} \{s_4\} & \{s_{4.9236}, s_{5.5485}, s_{5.5822}\} & \{s_{4.0706}, s_{5.4348}, s_{5.9647}\} \\ \{s_{3.0764}, s_{2.4515}, s_{2.4178}\} & \{s_4\} & \{s_{3.1518}, s_{3.8325}, s_{4.5011}\} \\ \{s_{3.9294}, s_{2.5652}, s_{2.0353}\} & \{s_{4.8482}, s_{4.1675}, s_{3.4989}\} & \{s_4\} \end{array} \right] \end{array}$$

**Step 4.** Calculate th worst consensus degree $wcd$ for the HFLPRs with respect to each index, and use the feedback mechanism in Algorithm II to modify the HFLPRs which are unsatisfied with the group consensus. Output the final outcomes with respect to each index, which is listed in Table VIII.

**Table VIII.** Performance evaluating outcomes of three funds with respect to each index

|  | Priorities of three funds | Ranking of funds |
|---|---|---|
| Index 1 | $w^{*,index1} = (0.3222, 0.4297, 0.2480)^T$ | $X_2 \succ X_1 \succ X_3$ |
| Index 2 | $w^{*,index2} = (0.4160, 0.2312, 0.3527)^T$ | $X_1 \succ X_3 \succ X_2$ |
| Index 3 | $w^{*,index3} = (0.4162, 0.3132, 0.2706)^T$ | $X_1 \succ X_2 \succ X_3$ |

According to the priority vector of the indexes policy efficiency, economic efficiency, and management efficiency: $\varpi = (0.3, 0.5, 0.2)^T$, we multiply this priority vector with three optimal priority vectors in Table VIII, then the final performance evaluating outcomes of the three funds are: $w^* = (0.3879, 0.3072, 0.3049)^T$, which indicates $X_1 \succ X_2 \succ X_3$. From the above outcomes, we can get that the second fund has more room to be improved at the aspect of economic efficiency. Furthermore, if the top management team pays more attention to the management performance of the third fund, it will bring more benefits to the enterprise

## 6. An application of the proposed algorithms--Online portal

Suppose that DMs or policymakers use the proposed algorithm to solve a GDM problem. Since redoing (programming) or deploying our work costs lots of effort and resources, and the essential background knowledge and programming skills are commonly limited, we introduce an online portal that involves two proposed algorithms as an application of our work that brings convenience for users.

The portal is publicly accessible at http://34.92.80.18/ (username: uniman; password: friendintegrityfaith). It provides a path for users to utilize our work to solve the decision-making problems without knowing the detail of the internal algorithms or background knowledge of the HFLPR.

Our online portal has the advantages: (1) zero to install: it only needs a browser to access; (2) centralized data, secure and easy to backup; (3) quick and easy to update, which are achieved by designing and deploying the online portal to a cloud server under the microservice architecture (Li, et al., 2018). The microservice architecture pattern quickly gains ground in the industry as a viable alternative to monolithic applications by arranging an application as a collection of loosely coupled services (Vresk & Čavrak, 2016). In other word, micorservices break an application into independent, loosely-coupled, individually deployable services. Containers are a lightweight, effieient and standard way for application to achieve this. Containers are a form of operating system virtualization. A container can be used to run everything from small microservices or software processes to large applications. The container contains all necessary executable files, binary code, libraries and configuration files. The management operations of the container including the creation, deployment and scaling of containers are named container management (Bernstein, 2014). The chief benefit of container management is the simplified management for clusters of container hosts. The fundamental unit of our designed microservice architecture is Pod (point of delivery). In terms of container management, a Pod is the basic execution unit of an application. In our case, the application (online portal) is designe to contain several GDM algorithms with HFLPR and each algoirithm is deployed to a Pod and run as a microservices, which means the online portal is compatible for other GDM algorithms with HFLPRs. These microservices form the decision support service of our online portal. Figure 5 is the overall architecture of our online portal. By using a container management tool, called Kubernetes (K8S), we can create a Pod pool that dynamically adds new algorithms, updates algorithms and removes algorithms. In addition, we create a web server to communicate with the uploaded algorithms in the Pod pool. Our proposed Algorithm I and II are already uploaded as a service in the online portal and accessible by the online users/DMs/scholars.

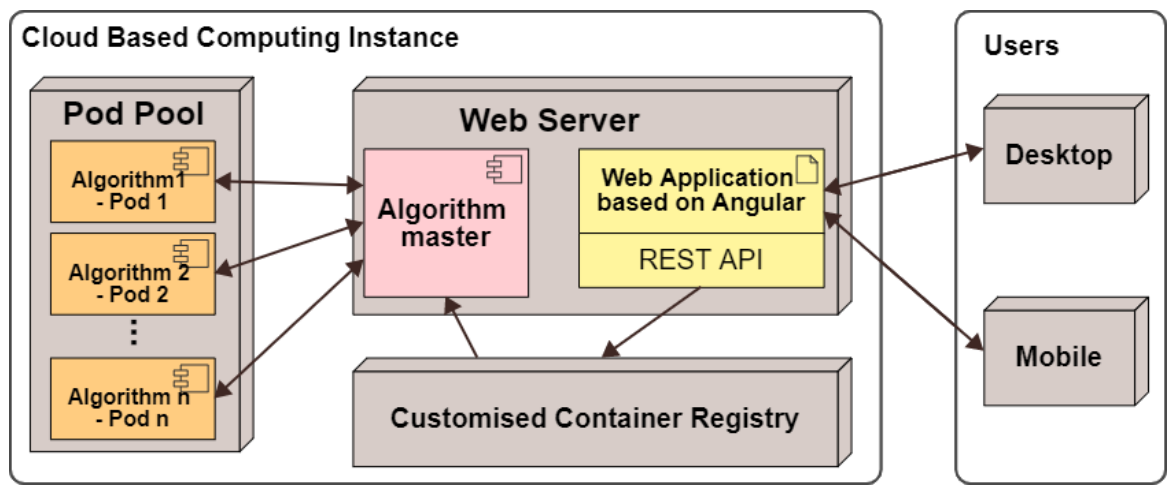

**Figure 5.** System architecture (Liu et al., 2021)

The UI of the portal is shown in Figure 6. The red box provides detailed explanations of how we transfer the HFLPRs to imputable formats for the convenience of users. Users could switch Algorithm 1 and Algorithm 2 in the "Algorithm Selection box". The number of alternatives could be specified in the box of "Number of Alternatives".

To enable user/DM to provide their HFLPRs based on the LTS $S = \{s_\alpha | \alpha = 0,1,...,2\tau\}$, we simplify the input form of the element in an HFLPR as shown in Figrue 6. The following rules which interpret the properties and definitions in Section 2.2 and 2.3 are created to restrict and validate user's input: (1) Users only need to input the subscript of each LT, which will make the calculation process easier in the back-end algorithms; (2) Users need to provide the maximum and minimum LTs rather than all LTs for each element of the HFLPR, where $HFLE(i,j)_{max} \geq HFLE(i,j)_{min}$; (3) All inputs must satisfy the property $HFLE(i,j)_{max} + HFLE(j,i)_{min} = s_{2\tau}$.

The above input rules are implemented by programming the customized validation functions to the input form. An error message will be shown under the corresponding input box if an user's inputs violates the second rule. The input form could prevent a user from inputting non-numeric characters with our coded validation function. Based on users' inputs, the support system will automatically recover all HFLEs according to the maximum and minimum LTs with considering other input restrictions, then send them to the back-end algorithms in the Pod pool.

For Algorithm I (Consistency procedure of a HFLPR), after a user inputs his/her HFLPR, the "submit" button will be activated if all inputs are correct with no errors reported by form validation functions. The inputted information is sent to the backend to process by clicking the "Submit" button. The result is shown in the box of "Final Result". Once the form is submitted, the user's original HFLPR will be shown in the box of "User's Input HFLPR" and at this stage, which can be checked for correctness.

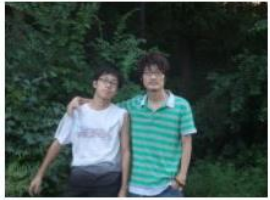

**Figure 6.** The portal UI

If users select Algorithm II (GDM based on the acceptable consistency and consensus measurements.) to solve the GDM problem, then the UI is changed to Figure 7. Figure 7 depicts the input of 4 experts' HFLPRs for three products under economic efficiency described in Section 5. Serval information boxes containing users' inputs are shown below. For the GDM problem, the users/DMs need to input the HFLPR one by one. The "submit" button would changes to "proceed" after inputting. The key information of the inputted HFLPRs is saved in the information box. Users can check information of the last inputted HFLPR through the box of "User's Input HFLPRs". The "clear" button could clear all information in the input form. The selection box above the input form "*consensus threshold value*" lets users select the consensus threshold value in the GDM process. Once finished, users could click the "Submit" button to get the ranking result, then the final alternative ranking and the number of rounds for Algorithm II to reach consensus would be returned to the webserver and shown in the box of "Final Result". The Figrue 8 shows the result of the final ranking of all three products and the number of rounds needed for Algorithm II to reach consensus under the input of priority vector of economic efficiency in Section 5.

A more detailed "User Guide" file which demonstrates how to implement this portal to solve decision-making problems can be downloaded by clicking the link of "click to download"

**Figure 7.** The UI of Algorithm II

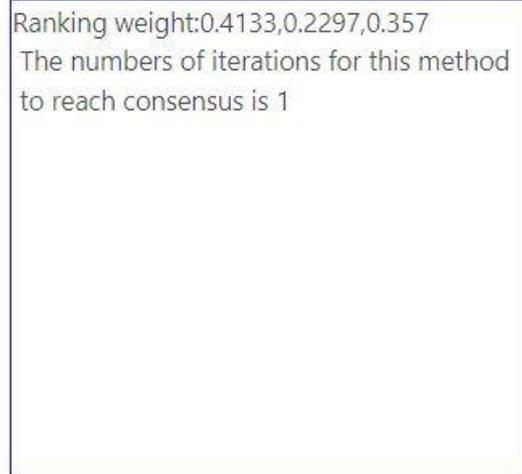

**Figure 8.** Example final result

## 7. Conclusions

As HFLPRs provide a more flexible and efficient expression way for DMs under uncertainty, the paper has investigated an algorithm for GDM with HFLPRs based on acceptable consistency and consensus measurements. The paper has discussed the relationship between the numerical and linguistic types of preference relations and has proposed the GCI for HFLPRs, called HFLGCI. Later on, the paper has measured the group consensus degree through the minimun similarity of each individual HFLPR and the overall perfect HFLPR, called worst consensus. A feedback algorithm has further been presented for reaching group consensus with acceptably consistent HFLPRs. Based on the proposed algorithm, the paper has designed experiments to discuss the critical values of HFLGCI with different orders of HFLPRs, and the critical values manifest the properties:

(1) For each order of HFLPRs, the critical value increases with the increase of $\alpha$, where $\alpha$ is a parameter in the algorithm to reconcile the difference between the priorities of any two alternatives;

(2) The changing trend of the critical value decreases with the increasing order of HFLPRs.

The paper has illustrated the practicability and effectiveness of the proposed algorithm by handling a case that evaluates the performance of venture capital guiding funds. Finally, an online decision-making portal is provided for DMs (users) to utilize the proposed algorithms to deal with decision-making problems.

For further work, there still exist great research prospect in the consistency and consensus theories of HFLPRs. For example, the perspectives, such as utilizing graph theory to investigate the consistency property, addressing the GDM with HFLPRs when DMs cannot compromise their evaluations, among others, have theoretical and practical significance.

**Acknowledgement**

The work was supported in part by the Guangdong Basic and Applied Basic Research Foundation (No. 2022A1515011029). We wish to thank Long Jin for his support and contribution to implementing the online portal.

**Appendix**

Part 1. Time and Space complexity analysis for Algorithm I

In each loop of Algorithm I, the inputted HFLPR get closer to the optimal solution. The HFLGCI will also be smaller than that in previous loops. Once $HFLGCI \leq \overline{HFLGCI}$, the algorithm stops. To analysis the time complexity of Algorithm I, the number of elementary operation of each step needs to be counted. The purpose of Step 2 is to calculate the priority vectors of the input HFLPR. Based on Proposition 3.1, the number of the priority vectors depends on the maximum $m_{ij}$ in the HFLPR where $m_{ij}$ is the number of LTs in $h_{ij}$. Since we set $\tau$ as 4, the granularity of the LTS, which is the maximum value of $m_{ij}$. Each priority vector contains $n$ elements. Thus, the number of elementary operation to calculate an element in the priority vectors is $n^2$ according to Equation (9). In total, the maximum number of elementary operations of step 2 is $9 \times n \times n^2$ and the time complexity is $O(n^3)$. From Equation (9), the number of calculations of Step 3 to get the HFLGCI from a priority vector is $(n^2 - n)/2$. Therefore, the time complexity is $O\left(\left(n^2 - n\right) \times 9/2\right) = O\left(n^2\right)$ since there are at most 9 vectors. In Step 5, the perfectly consistent LPR $\bar{L}$ is calculated by using at most $9n^2$ times of elementary operation, because its maximum size is $9n^2$ and $\bar{l}_{ij}$ can be obtained by using one step of the calculation based on Equation (9). The adjustment procedure for each element $b_{ij}^{o(\ell)}$ to derive a new HFLPR is the main factor to determines the speed of the consistency procedure and the number of loops in Algorithm I, where the adjustment function is proposed as $I(b_{ij}^{o(\ell)}(t+1)) = \beta I(b_{ij}^{o(\ell)}(t)) + (1-\beta)I(\bar{l}_{ij})$, where $0 \leq \beta \leq 1$. Normally, the value of $\beta$ is chosen from a range of 0.4 to 0.6. The maximum distance of the elements between an HFLPR and its perfectly consistent LPR is 9. Therefore, the maximum value of loop round in Algorithm 1 is $\log_{\left(\frac{1}{0.6}\right)} 9$. In conclusion, the time complexity of Algorithm I is $O\left(\log_{\left(\frac{1}{0.6}\right)} 9 \times \left(n^3 + 2n^2\right)\right) = O\left(n^3\right)$.

From the above analysis, we could see that the maximum units of memory space required to run Algorithm I is $3 \times 9 \times n^2 + 9 \times n + 10$, where $3 \times 9 \times n^2$ units of space are used for HLFPRs and LPRs, and $9 \times n$ units for the priority weight vectors, and 10 units for HFLGCIs. The space complexity of Algorithm I is $O\left(n^2\right)$.

Part 2. Time and Space complexity analysis for Algorithm II

Step 1 of Algorithm II is the implementation of Algorithm I for each DM, the time complexity is $O\left(n^3 k\right)$, where $k$ is the number of DMs. Based on the perfect consistent HFLPR calculation process proposed by Zhang and Chen (2019), the time complexity of obtaining the perfect ones for all DMs can be derived as $O\left(n^3 k\right)$. To get the weight of each DM, we need to calculate the similarity between each individual HFLPR and its perfect consistent HFLPR first. Since the granularity of the LTS is set as 9, the time complexity of similarity calculation can be deduced as $O\left(9n^2\right)$ according to the sinmilarity measure in (Liao et al., 2014). In total, the time complexity of Step 2 is $O\left(9kn^2 + n^3 k\right) = O\left(n^3 k\right)$. According to Equation (14), the maximum number of the elementary operations for proposing the collective perfectly consistent HFLPR is $9 \times k \times n^2$, therefore, the time complexity of Step 3 is $O\left(kn^2\right)$. As the consensus degree is defined as the minimum similarity between each individual HFLPR and the overall perfect HFLPR, the time complexity to calculated the consensus is is already analyzed in Step 2, which is $O(9kn^2)$. In Step 5, the modification function for an HFLPR is shown in Equation (16), which is similar to the adjustment function used in step 5 of Algorithm I. Like the loop analysis in Algorithm I, we can conclude that the maximum value of loop number in Algorithm II is $n \times k \times \log_{\left(\frac{1}{0.6}\right)} 9$, where $n \times k$ means the algorithm may modify all HFLPRs in the worst case. The revise function only has $n$ operations and the calculation of the consensus degree needs at most $n^2 \times 9$ operations. Therefore, the time complexity of the loop process in Algorithm II is $O\left(n \times k \times \left(\log_{\left(\frac{1}{0.6}\right)} 9\right) + \left(1 + n + 9n^2\right)\right) = O(n^3 k)$. For Step 6, the time complexity of obtaining the collective HFLPR and its priority weight vector was already analyzed in Step 3 of Algorithm II and Step 2 of Algorithm 1 respectively, both are $O(9kn^2)$. In conclusion, the time complexity of Algorithm II is $O(3n^3 k + 3n^2 k + 1) = O(3n^3 k)$.

During the calculation of Algorithm II, most of the memory space is used by all DMs' HFLPRs and their corresponding perfectly consistent HFLPRs that is $9 \times n^2 k \times 2$ units of memory space. The rest of memory is required by the collective perfectly consistent HFLPR ($9n^2$ units), weight vector of all DMs ($k$ units), and the priority weight vectors ($9n$ units). From this, we could conclude that the space complexity of Algorithm II is $O(n^2 k)$.